\pdfoutput=1

\documentclass[11pt]{article}

\usepackage{acl}

\usepackage{times}
\usepackage{latexsym}

\usepackage[T1]{fontenc}

\usepackage[utf8]{inputenc}

\usepackage{microtype}

\usepackage{inconsolata}

\usepackage{graphicx}
\usepackage{booktabs}
\usepackage{multirow}
\usepackage{xltabular}
\usepackage{enumitem}
\usepackage{paralist}
\usepackage{hyperref}
\usepackage{amsmath}

\title{Leveraging ChatGPT in Pharmacovigilance Event Extraction: An Empirical Study}

\author{
    Zhaoyue Sun\textsuperscript{\rm1}, 
    Gabriele Pergola\textsuperscript{\rm1},
    Byron C. Wallace\textsuperscript{\rm2}
\and
    Yulan He\textsuperscript{\rm1,3,4}\\
  \textsuperscript{1}Department of Computer Science, University of Warwick \\
  \textsuperscript{2}Khoury College of Computer Sciences, Northeastern University \\
  \textsuperscript{3}Department of Informatics, King's College London\\
  \textsuperscript{4}The Alan Turing Institute\\
  \texttt{\{Zhaoyue.Sun, Gabriele.Pergola.1\}@warwick.ac.uk} \\
  \texttt{b.wallace@northeastern.edu, yulan.he@kcl.ac.uk} \\
  }
  
\begin{document}
\maketitle
\begin{abstract}
With the advent of large language models (LLMs), there has been growing interest in exploring their potential for medical applications. This research aims to investigate the ability of LLMs, specifically ChatGPT, in the context of pharmacovigilance event extraction, of which the main goal is to identify and extract adverse events or potential therapeutic events from textual medical sources. We conduct extensive experiments to assess the performance of ChatGPT in the pharmacovigilance event extraction task, employing various prompts and demonstration selection strategies. The findings demonstrate that while ChatGPT demonstrates reasonable performance with appropriate demonstration selection strategies, it still falls short compared to fully fine-tuned small models. Additionally, we explore the potential of leveraging ChatGPT for data augmentation. However, our investigation reveals that the inclusion of synthesized data into fine-tuning may lead to a decrease in performance, possibly attributed to noise in the ChatGPT-generated labels. To mitigate this, we explore different filtering strategies and find that, with the proper approach, more stable performance can be achieved, although constant improvement remains elusive\footnote{Related code for this paper is available at \href{https://github.com/ZhaoyueSun/phee-with-chatgpt}{github.com/ZhaoyueSun/phee-with-chatgpt}.}.

\end{abstract}

\section{Introduction}

\begin{figure*}[h]
\centering
\includegraphics[scale=0.66]{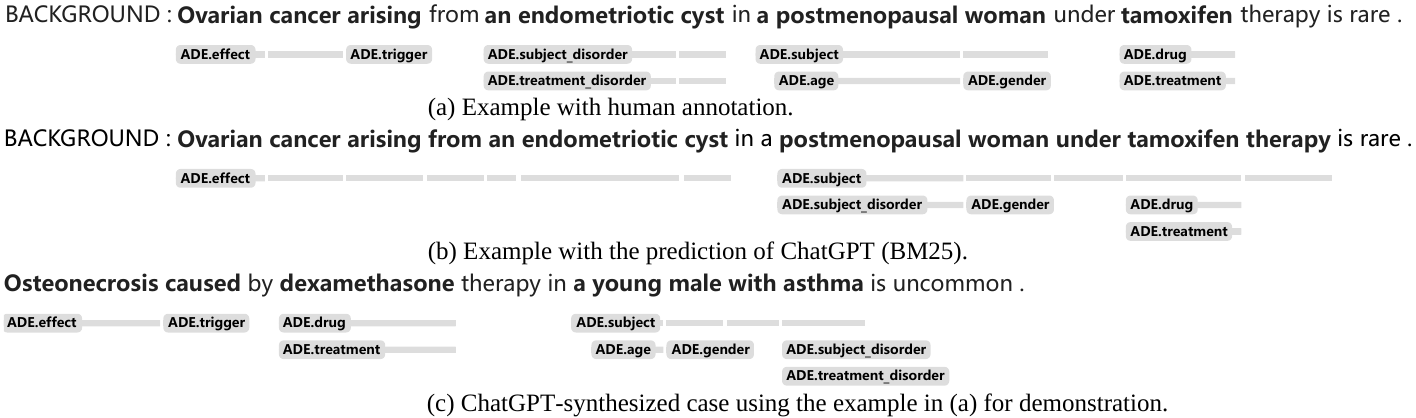} 
\caption{Snippets from biomedical documents: a comparison of human annotations, ChatGPT predictions, and a ChatGPT-synthesized case. \vspace{-12pt}}
\label{fig:example}
\end{figure*}

Pharmacovigilance stands as a pivotal discipline in healthcare that encompasses a range of processes: identifying, evaluating, understanding, and preventing adverse effects and other medicine-related issues \cite{WHO:04}. Within this domain, pharmacovigilance event extraction emerges as a crucial practice aimed at extracting structured medication-related event data from medical text sources, serving as valuable inputs for automatic drug safety signal detection. With the rapid expansion of electronic health records (EHR), medical case reports, and other textual resources, the need for efficient and accurate pharmacovigilance event extraction has become increasingly pressing.

Studies have been conducted to extract pharmacovigilance-related information from text data. However, previous research mainly focused on simple tasks such as entity extraction \cite{wunnava2017towards} or binary relation extraction \cite{gurulingappa:12,el2021mttlade}. Recently, \citet{sun-etal-2022-phee} introduced a novel dataset for pharmacovigilance event extraction, which includes hierarchical annotations of adverse events and potential therapeutic events, capturing information about the subject, treatment, and effect. Additionally, they investigate the performance of various models, including sequence labelling and QA-based approaches, for this task, providing a foundation for further advancements in extracting structured event data for pharmacovigilance research.

The rise of large language models (LLMs), especially ChatGPT \cite{chatgpt}, has sparked considerable interest in their potential applications in the medical field \cite{lu23napss, zhu23dis, agrawal2022large, kung2023performance}. In this study, our focus is on exploring different ways to incorporate ChatGPT into the pharmacovigilance event extraction task. Figure \ref{fig:example}(a) presents an example of this task. 

We first explore various strategies for prompting and demonstration selection to assess ChatGPT's performance in zero-shot and few-shot scenarios, comparing it with smaller fine-tuned models. Our findings indicate that, with suitable demonstration selecting strategies, ChatGPT performs reasonably well but still falls short of the performance achieved by fully fine-tuned smaller models, as demonstrated in Figure \ref{fig:example}(b). 

Furthermore, we delve into the utilization of LLMs for data augmentation, which is suggested to be beneficial in improving small model's performance in recent work \citep{boost21, lin2022few, liu2022wanli, zhu22dis, tan23event, whitehouse2023llm}. We employ ChatGPT to generate sentences structurally resembling demonstration samples, as illustrated in Figure \ref{fig:example}(c). However, our experiments show that simply combining these generated samples with the training set leads to an overall performance decrease. Considering the possible influence of synthesized data noises, we further introduce a filtering strategy for augmented data quality control, which, though still does not outperform the fully finetuned model, reduces the performance drop and brings it closer to the levels achieved with the original training data, while reducing the variance. This indicates enhanced stability when working with ample high-quality data. 

In summary, we compare various regimes of leveraging ChatGPT to assist in pharmacovigilance event extraction, providing practitioners with meaningful references for choosing suitable strategies. Additionally, we conduct a fine-grained qualitative analysis of ChatGPT synthesized instances and data augmentation and explore reasons for their lack of positive effect, laying the groundwork for improvements in subsequent work.





\section{Prompt-based Learning with ChatGPT}

\subsection{Zero-shot Prompting} 
For zero-shot prompting, a manually designed instruction is employed to query ChatGPT for answers. In this study, we devise four approaches to prompt the model: \begin{inparaenum}[a)] 
\item\textbf{Schema:} providing instructions alongside enumeration of event types and argument types;
\item\textbf{Explanation:} providing instructions with a detailed explanation of the schema;
\item\textbf{Code:} formulating instructions and output format using a combination of text descriptions and code snippets;
\item\textbf{Pipeline:} querying the model in a pipeline manner, which first prompts for the main arguments and then follows up with type-related questions for each sub-argument.
\end{inparaenum} Details of the prompts are presented in Appendix \ref{apd:prompt}.

\subsection{Few-shot In-context Learning} 
For few-shot in-context learning, several demonstrations are provided together with the instruction. The selection of different demonstration examples can yield varying results. We explore different strategies for choosing in-context examples based on a given test instance, including: \begin{inparaenum}[a)] 
\item\textbf{Random:} randomly selecting examples from the training set;
\item\textbf{SBERT:} choosing examples based on the similarity of their dense representations to the test sentence. We utilize Sentence-BERT \cite{reimers2019sentence} to obtain the sentence representations;
\item\textbf{BM25:} selecting examples based on the similarity of their lexical representations to the test sentence. We employ BM25 \cite{trotman2014improvements} as the ranking function;
\item\textbf{TreeKernel:} choosing examples based on the structural similarity to the test sentence. We implement the tree kernel by computing the Jaccard similarity of the subpaths within the dependency trees of the sentences.
\end{inparaenum}


\section{ChatGPT as Data Synthesizer}

We explore the potential of leveraging ChatGPT for data augmentation purposes. To achieve this, we incorporate an example from the training set, along with its annotated events, as input to ChatGPT. We then prompt ChatGPT to generate a sentence that exhibits a similar event structure to the given sentence and extract the events from the generated sentence. However, based on our initial study, we observed that ChatGPT tends to miss specific mentions of drugs or excessively use certain drugs, such as `\emph{ibuprofen}'. We address this issue by restricting the inclusion of drug names and their corresponding effects sampled from the training data in generated sentences. Details of the prompt for data synthesizing are shown in Appendix \ref{apd:prompt}.



Recognizing that directly incorporating generated samples into the training data can lead to performance decline, possibly due to issues related to data quality, we have introduced filtering strategies. The main rationale behind the filtering is to retain annotations for which the model exhibits a certain level of confidence, based on the assumption that a finetuned model possesses some discriminatory ability regarding the quality of annotations, and incorrect annotations may result in lower confidence scores from the model for the annotation sequence. Specifically, we introduce: 
\begin{inparaenum}[a)] 
\item\textbf{Train Filter:} Filtering the training set with $s_{gold} < \text{mean}(s_{gold})$, where $s_{gold}$ is the average token probability given by the fine-tuned model on the ground-truth event label sequence. This means we filter out training instances for which the annotation sequence has less model certainty than the average level;
\item\textbf{Augment Filter:} Filtering augmented data with  $z(s_{gold})<0$ or $z(s_{gold})<z(s_{pred})$, where $s_{pred}$ is the average token probability for predicted event label sequences. $z(s) = (s - \text{mean}(s^v))/\text{std}(s^v)$, and $s^v$ represents the values of $s$ in the validation set. In this case, we filter out samples generated by ChatGPT if their generated annotation sequence probability, as calculated by a fine-tuned model, falls below the average level or is less certain than the sequence predicted by the fine-tuned model itself. Considering models potentially assigning high scores to the sequences they predict, we use z-score instead of direct predictive probabilities for comparison.

\end{inparaenum} 

With these filtering rules, we compare the model's performance on several data settings, including: training data (Tr.), training data combined with augmented data (Tr.+Aug.), filtered training data (Tr. Fil.), training data with filtered augmented data (Tr.+Aug. Fil.) and filtered training data with filtered augmented data (Tr. Fil.+Aug. Fil.).

\section{Experiments}
\label{sec:exp}

\subsection{Experimental Settings}
\paragraph{Dataset} We conducted experiments on the PHEE dataset \cite{sun-etal-2022-phee}, 
an English event extraction dataset sourced from publicly accessible medical reports, encompassing annotations for two event categories: \emph{adverse events} and \emph{potential therapeutic events}. The annotations follow a hierarchical structure, with main arguments providing information on the \emph{subject}, \emph{treatment}, and \emph{effect}, while sub-arguments offer more detailed information pertaining to the main arguments. However, during our analysis, we observed that certain argument types showed low consistency. To address this issue, we performed automatic and manual revisions on the \emph{subject.disorder}, \emph{time\_elapsed}, and \emph{duration} arguments. For further details, please refer to Appendix \ref{apd:revision}. The dataset contains around 5k sentences and we split the data into training, validation, and test sets by 6/2/2. 

\paragraph{Baselines} We compare ChatGPT's performance with the best-performing Generative QA model proposed in \cite{sun-etal-2022-phee} and two widely adopted seq-to-seq models: 1) \textbf{UIE} \cite{lu-etal-2022-unified}, a model that is specifically pre-trained on structured information extraction data; and 2) \textbf{Flan-T5} \cite{chung2022scaling}, a model trained on a diverse range of tasks using instructional prompts. For more information, see Appendix \ref{apd:baseline}.

\paragraph{Evaluation} We follow \citet{sun-etal-2022-phee} to evaluate both exact matching F1 score (EM\_F1) and token-level matching F1 score (Token\_F1) for argument extraction. During our preliminary experiments, we observed that ChatGPT struggled to generate reasonable results for trigger extraction. Considering that even humans find trigger identification challenging, and that it doesn't significantly contribute to understanding pharmacovigilance events, we did not query ChatGPT for triggers, but we still ask ChatGPT to generate the event structure, enabling the differentiation of multiple events. For the trigger extraction results obtained from finetuning models, please check Appendix \ref{apd:trigger-result}.

We perform 5-fold cross-validation for fine-tuning and data augmentation experiments, while limiting ChatGPT-based zero-shot and few-shot learning to a single split due to cost-related reasons. For more details about the experimental setup, please refer to Appendix \ref{apd:exp-setup}.

\subsection{Results and Discussion}
\label{sec:discussion}

\paragraph{ChatGPT with Different Prompting Strategies}

\begin{table}[h]
\small
\centering
\resizebox{0.97\columnwidth}{!}{
\begin{tabular}{@{}lcccc@{}}
\toprule
  & \multicolumn{2}{c}{Main-arguments} & \multicolumn{2}{c}{Sub-arguments} \\ \cmidrule(l){2-3} \cmidrule(l){4-5} 
   & EM\_F1 & Token\_F1 & EM\_F1 & Token\_F1 \\ \midrule
Schema       & 30.31          & 47.41             & 22.50          & 26.51            \\
Code       & 25.94          & 40.42             & 25.67          & 29.70            \\
Explanation      & \textbf{34.80}          & \textbf{52.99}             & \textbf{36.70}          & \textbf{39.33}            \\
Pipeline       & 32.57          & 49.41             & 27.79          & 33.80            \\ \hline

\end{tabular}}
\caption{Argument extraction results for ChatGPT zero-shot prompting with different prompting strategies.}

\label{tab:prompt-result}
\end{table}

Table \ref{tab:prompt-result} presents the argument extraction results for ChatGPT using different zero-shot prompting strategies. Providing only instructions yields unsatisfactory performance, but including a detailed explanation of the event schema leads to noticeable improvement, highlighting the importance of comprehensive guidance. Further qualitative examination reveals that end-to-end generation tends to miss arguments, whereas the pipeline approach tends to generate numerous false positive cases. It is surprising that the model performs poorly on seemingly simple arguments such as `\emph{population}', `\emph{route}', and `\emph{age}'. While providing explanations improves the performance of some arguments (e.g., `\emph{route}' and `\emph{age}'), all approaches still struggle with `\emph{population}' extraction. This difficulty may due to the gap between the lexical meaning of the label `\emph{population}' and the semantic meaning of the argument. Additionally, while the pipeline method has advantages in extracting certain argument types (e.g., `\emph{gender}' and `\emph{frequency}'), the inference time is proportional to the number of argument types, making it approximately 10 times longer than the end-to-end methods.

Table \ref{tab:demo-result} displays the few-shot argument extraction results for ChatGPT using various in-context selection strategies. Dense representation-based demonstration retrieval with SBERT does not demonstrate superiority in this task, possibly due to limited domain knowledge captured by the pre-trained sentence representation model. Incorporating structured information improves performance, while the simplest lexical-based retrieval strategy shows the most noticeable performance gains. Upon examining the samples retrieved by different example selection strategies, we observed that SBERT and TreeKernel tend to retrieve structurally similar sentences, while BM25 is more inclined to retrieve sentences containing matching entities such as drugs (since entities usually serve as keywords in a sentence). This observation suggests that the superior performance of BM25 in argument extraction can be attributed to the fact that this task is more sensitive to entities. When more examples with similar entities are covered, ChatGPT learns more effectively from them.

\begin{table}[h]
\small
\centering
\resizebox{0.97\columnwidth}{!}{
\begin{tabular}{@{}lcccc@{}}
\toprule
  & \multicolumn{2}{c}{Main-arguments} & \multicolumn{2}{c}{Sub-arguments} \\ \cmidrule(l){2-3} \cmidrule(l){4-5} 
   & EM\_F1 & Token\_F1 & EM\_F1 & Token\_F1 \\ \midrule
random       & 58.31          & 72.74             & 60.32          & 63.74            \\
SBERT       & 56.90          & 71.65             & 62.29          & 64.25            \\
TreeKernel       & \textbf{60.54}          & 73.68             & 63.36          & 64.69            \\
BM25      & 60.39          & \textbf{76.15}             & \textbf{67.35}          & \textbf{68.67}            \\\hline

\end{tabular}}
\caption{Argument extraction results for ChatGPT few-shot prompting with different in-context demonstration selection strategies (results for 5-shot are reported). }

\label{tab:demo-result}
\end{table}

\paragraph{Finetuning Models vs. ChatGPT}

Table \ref{tab:main-result} illustrates the argument extraction results for different methods. The findings indicate that there is minimal variation among the fine-tuning methods. Specifically, the Flan-T5 model, despite not being explicitly pre-trained for the information extraction task, demonstrates slightly better performance than the UIE model. In contrast, ChatGPT without demonstrations exhibits poor performance. However, when demonstrations are provided, ChatGPT shows improved results, although there remains a noticeable gap compared to the fine-tuning methods. For a detailed breakdown of the results for each argument type, refer to Appendix \ref{apd:arg-result}. 

\begin{table}[h]
\centering
\resizebox{0.98\columnwidth}{!}{
\begin{tabular}{@{}lcccc@{}}
\toprule
  & \multicolumn{2}{c}{Main-arguments} & \multicolumn{2}{c}{Sub-arguments} \\ \cmidrule(l){2-3} \cmidrule(l){4-5} 
   & EM\_F1 & Token\_F1 & EM\_F1 & Token\_F1 \\ \midrule
\multicolumn{5}{c}{Fully supervised}                                                                \\ \hline
Generative QA 
& 68.85          & 81.63             & 77.33          & 78.83            \\
UIE(Large)           & $69.46_{\pm.49}$          & $81.20_{\pm.40}$             & $77.12_{\pm1.3}$          & $78.83_{\pm1.4}$            \\
Flan-T5(Large)       & $\textbf{70.78}_{\pm1.4}$         & $\textbf{82.34}_{\pm1.5}$             & $\textbf{77.63}_{\pm.1.6}$          & $\textbf{79.52}_{\pm1.3}$            \\ 
\hline
\multicolumn{5}{c}{Zero-Shot}                                                                 \\ \hline
ChatGPT(Exp.) & 34.80          & 52.99             & 36.70          & 39.33            \\ \hline
\multicolumn{5}{c}{Few-Shot}                                                                  \\ \hline
ChatGPT(BM25)        & 60.39          & 76.15             & 67.35          & 68.67            \\ \hline
\end{tabular}}
\caption{Argument extraction results for various methods. For fine-tuning methods, we report the $mean_{\pm std}$ value of 5-fold cross-validation. For ChatGPT(BM25), we provide the results for 5-shot. We obtain Generative QA results directly from the original paper.}

\label{tab:main-result}
\end{table}

\paragraph{Data Augmentation with ChatGPT}

Table \ref{tab:aug-result} presents the performance of Flan-T5 when augmentation and various filtering strategies are employed. It can be seen that simply extending the training data with ChatGPT-synthesized cases could lead to an obvious performance drop. In contrast, with the filtered training set, although retained only 65\% of training data, surpasses results obtained from over 5,000 augmented instances, which may indicate the critical role of data quality in pharmacovigilance event extraction. Furthermore, training with filtered augmented data effectively restores performance to the original level. In particular, training with both filtered training data and filtered augmented data displays only slight deviations from training with the original data, yet it reduces variance. The p-values for the variance difference significance, assessed through the F-test, are 0.29 and 0.39 for EM\_F1 and Token\_F1, respectively.

\begin{table}[h]
\centering
\resizebox{0.97\columnwidth}{!}{
\begin{tabular}{@{}lccc} 
\toprule
   & EM\_F1 & Token\_F1 & Avg. Cases\\ \midrule
Tr.& $\textbf{74.45}_{\pm1.46}$& $\textbf{81.30}_{\pm1.27}$& 2897\\ 
 Tr.+Aug.& $73.07_{\pm0.92}$&  $79.93_{\pm1.51}$& 5446\\\hline
 Tr. Fil.& $73.92_{\pm.1.28}$& $80.71_{\pm1.60}$& 1873\\
 Tr.+Aug. Fil.& $74.26_{\pm.1.27}$& $80.98_{\pm2.06}$& 3702\\
 Tr. Fil.+Aug. Fil.& $74.19_{\pm1.09}$& $81.05_{\pm1.09}$& 2678\\\hline

\end{tabular}}
\caption{Argument (including main and sub-arguments) extraction results for Flan-T5 (Large) with augmentation and filtering strategies. The \textit{Avg. Cases} column displays the average number of training cases over 5 folds.}

\label{tab:aug-result}
\end{table}


We conduct a qualitative analysis to explore possible reasons for the performance degradation caused by data augmentation. Through sampling analysis of examples where the fine-tuned model made correct predictions but the augmented model failed, we find that for main arguments, the errors mainly stemmed from inconsistency in text span boundaries, while failures due to semantic misunderstandings are relatively rare and primarily occurred in the misidentification of abbreviations, such as the model incorrectly recognizing an abbreviation for a disease as a medication. As for sub-arguments, semantic misunderstandings and missing arguments are the main reasons the augmented model makes mistakes.
Additionally, some errors resulted from inconsistent boundaries and annotation noise, which may influence evaluation scores but not necessarily harm model utility. 

Cases where the arguments are missing sometimes show a pattern, e.g., a `\textit{population}' argument tends to be missed when it’s in an expression like `xx cases’, and a `\textit{route}' argument may be missed when overlapping with the `\textit{dosage}' arguments. However, in many cases, there’s no obvious reason why the argument is not extracted. For semantic issues, we observe that the `\textit{subject.disorder}' is easily confused with `\textit{treatment.disorder}', and  `\textit{time elapsed}' is easily confused with `\textit{duration}'. Additionally, some `\textit{age}’ expressions (e.g., `adults') tend to be predicted as `\textit{gender}', and some `\textit{duration}' expressions when describing a long term may be identified as `\textit{frequency}'. 

Furthermore, we conduct an unconditional sampling of instances synthesized by ChatGPT and analyze the mislabels. The analysis indicates that, although category labelling errors are not common in ChatGPT-synthesized samples, there are still instances of mislabelling for some relatively challenging arguments, such as `\textit{subject.disorder}' and `\textit{treatment.disorder}'. Moreover, we observe that, in comparison to the types and quantities of arguments present in the given templates, ChatGPT-synthesized examples have less coverage for rare arguments. This might be one of the reasons contributing to performance degradation in argument extraction when using synthetic data for augmentation as well. 

According to the qualitative analysis, we suspect the model may struggle to capture the intricate annotation rules when only one example is used as a demonstration. For future work, providing more diverse examples when synthesizing data may be a worthwhile direction to explore. For more details on qualitative analysis, please refer to Appendix \ref{apd:qualitative-analysis}.

\section{Conclusion}

This paper provides empirical practice in various approaches to leveraging ChatGPT for the pharmacovigilance event extraction task. Overall, ChatGPT exhibits impressive few-shot learning capabilities in pharmacovigilance event extraction. Nevertheless, considering the sensitivity of the medical field, fine-tuned models retain a clear edge 
in the presence of abundant data. 
In our experiments, the introduction of ChatGPT-synthesized instances for data augmentation does not improve the performance of small model fine-tuning. However, appropriate quality control may increase the stability of performance. Qualitative analysis indicates that errors may arise in ChatGPT-synthesized data when distinguishing semantically complex arguments, and the coverage of rare arguments is insufficient. We emphasize the structural complexity and fine granularity of arguments in event extraction, which may pose challenges in generating synthetic data. Future work can conduct more in-depth data augmentation research addressing these aspects.

\section*{Limitations}
In our preliminary study, we encountered limitations in exploring alternative open-source LLMs, such as LLaMA 30B \cite{touvron2023llama} and Flan-T5 XXL \cite{chung2022scaling}, for zero-shot/few-shot prompting. These models exhibited significant differences in generation quality compared to ChatGPT, and their slow inference speeds hindered a comprehensive evaluation. Despite these limitations, we highlight the importance of further research to investigate the potential of leveraging different open-source LLMs.

Secondly, while Chain-of-Thought (CoT) reasoning has demonstrated enhanced performance in few-shot learning for biomedical NLP tasks, we have not yet introduced it into our evaluation. This omission is attributed to the intricate nature of constructing reasoning steps for each argument within the fine-grained event extraction task. In our preliminary experiments, simply asking ChatGPT to explain its extraction rationale didn't enhance performance; instead, it complicated the accurate collection of extraction results. In our current experiment, we explored ways to retrieve the most relevant examples from the training set, but we lacked annotated reasoning steps for all samples, hindering a comprehensive evaluation of the CoT method in this context. Given these limitations, we leave the exploration of CoT to future work.

Moreover, our investigation focused solely on unsupervised methods for in-context demonstration selection. Future research could explore the incorporation of annotations in the selection process, which may yield valuable insights and improve the performance of ChatGPT in pharmacovigilance event extraction.

\section*{Ethics Statement}

The approaches outlined in this article focus solely on extracting information from the textual level and do not suggest a direct causal relationship between drugs and their effects. The causality assessment of ADEs requires expert evaluation, and the methodologies presented in this paper are intended as supplementary tools to accelerate the process.

This paper explores common errors inherent in the model's extraction, and users should be aware of the practical consequences associated with different error types. Caution is particularly advised when employing statistical inferences based on the tools proposed in this paper, as the model may sometimes miss an argument (such as failure to recognize a patient's race in the case of certain place names due to insufficient generalization). Additionally, in many instances, the extracted sentences themselves may not mention certain pieces of information. In comparison to free-text sources, structured data such as EHRs may offer a more reliable basis for conducting statistical inferences.

Furthermore, it is worth noting that the use of ChatGPT-synthesized data may alter the data distribution. For example, we observe that ChatGPT is more likely to generate the most common drug-ADE pairs. Although this is reasonable, including a substantial number of such "correct" examples in the training data may lead to model bias, causing it to overlook rare but significant side effects mentioned in the text. Data synthesized by ChatGPT may also introduce incorrect knowledge, while the impact of this on event extraction may be limited because the given sentence constrains the extraction results. However, caution is needed when applying the methods described in the text to other application domains, such as ADE generation.

\section*{Acknowledgements}

This work was supported in part by the UK Engineering and Physical Sciences Research Council through a Turing AI Fellowship (EP/V020579/1, EP/V020579/2) and the National Science Foundation (NSF) grant 1750978. 

\bibliography{anthology,custom}

\newpage

\appendix

\setcounter{table}{0}
\renewcommand{\thetable}{A\arabic{table}}
\setcounter{figure}{0}
\renewcommand{\thefigure}{A\arabic{figure}}

\section{Data Annotation Revision Details}
\label{apd:revision}

In the original dataset, we observed particularly low levels of annotation inconsistency for `\emph{subject.disorder}', `\emph{time\_elapsed}', and `\emph{duration}' arguments, as illustrated by the examples provided in Table \ref{tab:data-revision}. To address this, we conducted an automatic revision for the `\emph{subject.disorder}' annotation and hired annotators to manually correct the `\emph{time\_elapsed}' and `\emph{duration}' annotations. For `\emph{subject.disorder}' correction, if a `\emph{treatment.disorder}' was present in the `\emph{subject}' argument but not annotated as `\emph{subject.disorder}', we added it to the `\emph{subject.disorder}' annotation. For `\emph{time\_elapsed}' and `\emph{duration}' correction, detailed guidelines were provided to the annotators to ensure consistent annotations. We employed three annotators and informed them about the purpose of the data. The annotators are all PhD students who volunteered for this task, receiving compensation through the university's payment platform for their annotation work. Two of the annotators have a background in computer science, and one annotator has a medical background. Two of the annotators are non-native English speakers, and one is a native English speaker. Following the approach used in \cite{sun-etal-2022-phee}, we evaluated the consistency among the annotators using the EM\_F1 score. The averaged EM\_F1 scores for both `\emph{time\_elapsed}' and `\emph{duration}' annotations were 75.3\%.

\begin{table*}[tbh]
\small
\begin{tabular}{p{0.95\textwidth}}
\toprule
\textbf{Subject.Disorder: } We report two patients with {\color{red}acne vulgaris} with a fourth type of minocycline-induced cutaneous pigmentation.\\
We observed that when a disorder span is included in a `subject' argument and also as `treatment.disorder', annotations in the original dataset show inconsistency on whether to annotate this span as `subject.disorder'.\\
\hline
\textbf{Time\_elapsed \& Duration: } \\
In this article, we describe another case of subcutaneous changes following repeated glatiramer acetate injection, presented as localized panniculitis in the area around the injection sites, in a 46-year-old female patient who was treated with glatiramer acetate for {\color{red}18 months}. \\
Annotation inconsistencies arise when a `time\_elapsed' argument can also be described as `duration'.\\
\bottomrule
\end{tabular}
\caption{Inconsistent examples from the PHEE dataset.}
\label{tab:data-revision}
\end{table*}

\section{Details of Baseline Implementation}
\label{apd:baseline}

For the implementation of seq-to-seq baselines, we formulate pharmacovigilance event extraction as a conditional text generation task. Concretely, given a sentence $x$ and additional auxiliary information $a$, the model is trained to generate a linearized sequence $y$ representing the output event structure.

For UIE, we refer to the methodology outlined in the original paper by utilizing the Structural Schema Instructor (SSI) as the auxiliary information $a$ and constructing the target sequence $y$ with Structural Extraction Language (SEL). However, special tokens used in SSI and SEL in UIE can result in a decrease in performance if no external pre-training is applied. Thus for Flan-T5, we substitute the SSI with a concise instruction accompanied by a natural language enumeration of the schema. Additionally, for the target sequence construction, we utilize square brackets as the structural symbol.

For both UIE and FLan-T5, we use the large model which comprises 770M parameters. Training an epoch typically takes around 2 minutes, and validation, which utilizes beam search, requires approximately 10 minutes with an NVIDIA A100 (80G) GPU. The fine-tuning models generally converge within 10 epochs.

\section{Details of Experimental Setup}
\label{apd:exp-setup}
\subsection{Few-shot Prompting Settings}

In the context of event extraction, each shot includes one example for each event type. In Section \ref{sec:exp}, we report the 5-shot results for in-context demonstration selection strategies, which entails providing a total of 10 examples for each instance. The selection of the number of demonstration cases was based on ChatGPT's input length capacity. 

\begin{figure}[h]
\centering
\includegraphics[scale=0.22]{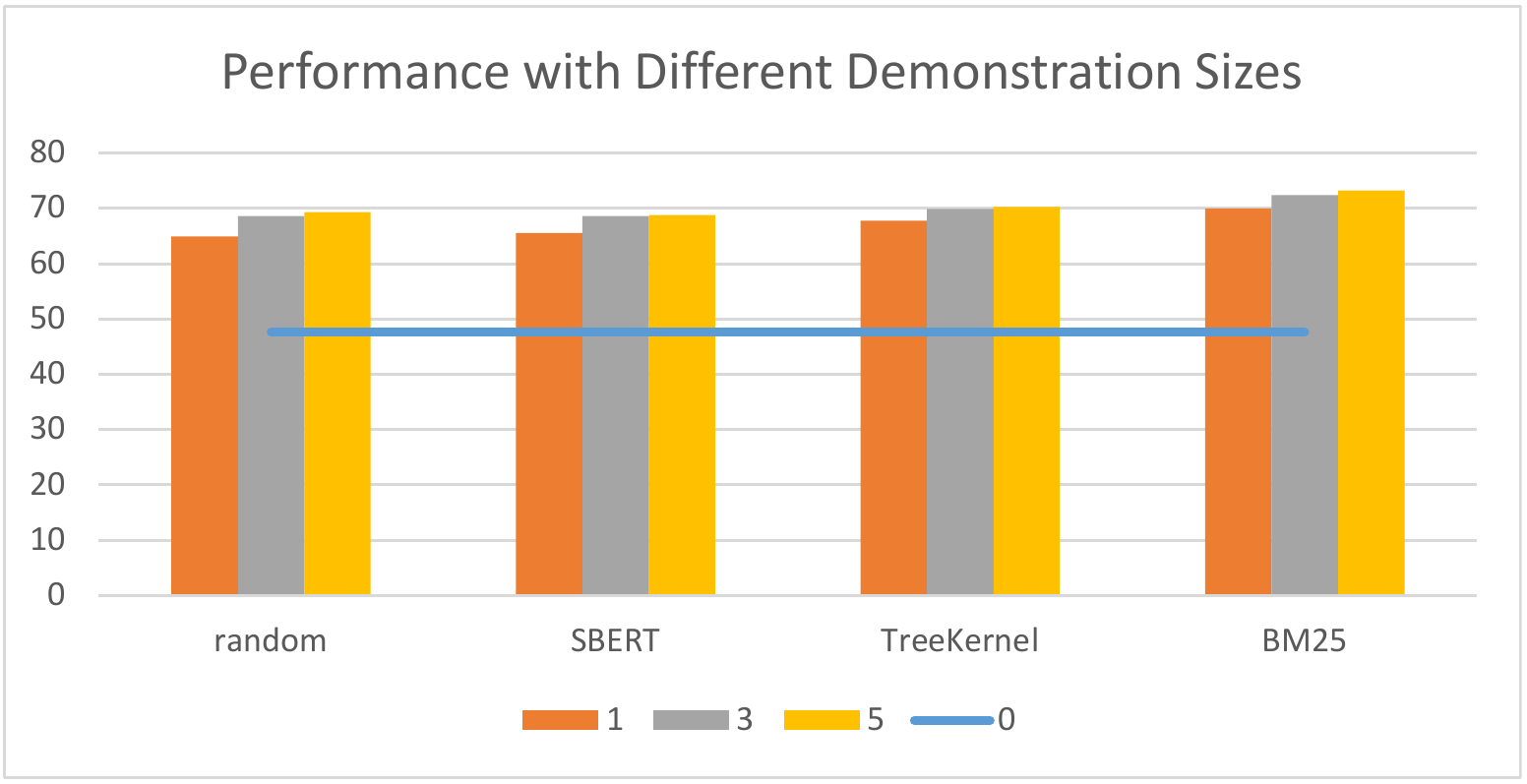} 
\caption{Token\_F1 scores for argument extraction with different demonstration sizes. The blue line represents the performance of zero-shot prompting with the explained schema.}
\label{fig:demo_size}
\end{figure}

We further evaluate the argument extraction performance of several in-context demonstration selection strategies when different numbers of demonstration examples are selected in Figure \ref{fig:demo_size}. Notably, when the first example is added, all methods experience a significant performance boost. However, as the number of examples increases, the performance gains become more minimal. Five-shot prompting (involving 5 ADE examples and 5 PTE examples) has approached the maximum input limit that ChatGPT can handle. Nevertheless, we reasonably suspect that further increasing the number of examples would not get significant performance improvements.


\subsection{Hyperparameter Details}
\label{apd:hyperparam}

The order and occurrence of events and arguments in the generated sequence can impact the learning effectiveness of the model. To tackle this, \citet{lu-etal-2022-unified} introduced the `Rejection Mechanism', which generates a null span when a specific type of event or argument is absent in the sentence. In our preliminary experiments, we determined that the noise injection ratio has little impact on the performance but the order of the argument generation matters. Therefore, we choose to set the noise injection ratio to 0 and keep the arguments generated in order to reduce the fluctuation caused by random insertion during model comparison.



To fine-tune the models, we establish a maximum length of 512 tokens for both input and output. We utilize a total batch size of 32 for the large model, and 64 for the base model. The learning rates are configured as 3e-4 for the large model and 5e-4 for the base model, with a warm-up ratio of 0.06. We train the models for a maximum of 50 epochs, early stopping if there is no improvement for 5 epochs. During the generation process, we employ beam search with a beam size of 3. 

We employ the `gpt-3.5-turbo-0301' version of ChatGPT for prompting-based event extraction and synthesized data generation. The temperature is set as 0 for zero-shot and few-shot prompting, and 0.2 for data generation. 



\section{Trigger Extraction Results for Finetuning Methods}
\label{apd:trigger-result}

Table \ref{tab:trigger} displays the results of trigger extraction and event type classification for the fine-tuning models. In general, there is little difference in the performance of trigger extraction and event type classification between different models. Furthermore, training with filtered training and augmented data still exhibits the smallest variance, which is consistent with the observation for argument extraction.


\begin{table}[h]
\centering
\resizebox{\columnwidth}{!}{
\begin{tabular}{@{}lcc@{}}
\toprule
 & Trigger & Event Type \\ \midrule
UIE(Large) & $\textbf{69.92
}_{\pm1.72}$& $94.78
_{\pm.72}$\\\hline
Flan-T5(Large) & $69.60_{\pm1.87}$& $95.04_{\pm.97}$\\
 w/ Tr.+Aug.& $68.46_{\pm1.83}$&$94.92_{\pm.60}$\\
 w/ Tr. Fil.& $69.50_{\pm1.61}$&$94.92_{\pm.88}$\\
 w/ Tr.+Aug. Fil.&$69.68_{\pm1.36}$&$95.00_{\pm.79}$\\ 
w/ Tr. Fil.+Aug. Fil.& $69.73_{\pm1.14}$  & $\textbf{95.13}_{\pm.48}$  \\ \bottomrule
\end{tabular}}
\caption{Results for trigger extraction (EM\_F1) and event type classification (F1).}
\label{tab:trigger}
\end{table}

\section{Argument Extraction Results for Each Argument Type}
\label{apd:arg-result}

\begin{table*}[htb]
\centering
\resizebox{0.97\textwidth}{!}{
\begin{tabular}{lllllllcc}
\hline
 &
  \multicolumn{2}{c}{Flan-T5} &
  \multicolumn{2}{c}{\begin{tabular}[c]{@{}c@{}}Flan-T5\\ (Tr.+Aug.)\end{tabular}} &
  \multicolumn{2}{l}{\begin{tabular}[c]{@{}c@{}}Flan-T5\\ (Tr. Fil.+Aug. Fil.)\end{tabular}} &
  \multicolumn{2}{c}{ChatGPT} \\ \cline{2-9} 
 &
  EM\_F1 &
  \multicolumn{1}{c}{Token\_F1} &
  \multicolumn{1}{c}{EM\_F1} &
  \multicolumn{1}{c}{Token\_F1} &
  \multicolumn{1}{c}{EM\_F1} &
  \multicolumn{1}{c}{Token\_F1} &
  EM\_F1 &
  Token\_F1 \\ \hline
Subject                                     & \textbf{73.11} & \textbf{82.37} & 70.93 & 80.90 & 72.39 & 82.15 & 57.96 & 75.20 \\
\textit{ Age}              & \textbf{88.12} & 92.07 & 87.21 & 92.55 & 87.50 & \textbf{92.82} & 86.62 & 90.18 \\
\textit{ Disorder}         & \textbf{69.80} & 77.13 & 63.81 & 72.76 & 69.73 & \textbf{77.45} & 53.90 & 61.08 \\
\textit{ Gender}           & 86.73 & 86.51 & 86.03 & 85.78 & \textbf{87.15} & \textbf{87.00} & 84.29 & 85.07 \\
\textit{ Population}       & 74.83 & 75.72 & 72.30 & 73.94 & \textbf{75.90} & \textbf{76.69} & 49.30 & 42.11 \\
\textit{ Race}             & 93.20 & \textbf{93.35} & \textbf{93.29} & 91.20 & 92.02 & 91.52 & 87.5  & 77.78 \\ \hline
Treatment                                   & \textbf{66.35} & \textbf{79.82} & 66.27 & 79.00 & 65.90 & 79.68 & 57.67 & 73.49 \\
\textit{ Drug}             & \textbf{87.03} & \textbf{88.32} & 85.84 & 87.45 & 86.65 & 87.99 & 80.78 & 82.59 \\
\textit{ Disorder}         & \textbf{67.19} & \textbf{73.14} & 65.24 & 71.73 & 66.64 & 72.57 & 55.89 & 62.01 \\
\textit{ Route}            & \textbf{67.76} & 69.34 & 63.55 & 65.55 & 66.37 & \textbf{70.39} & 56.66 & 63.73 \\
\textit{ Dosage}           & \textbf{65.95} & \textbf{76.40} & 63.58 & 72.17 & 62.91 & 73.16 & 47.11 & 61.05 \\
\textit{ Time elapsed}     & 61.56 & 71.21 & 54.11 & 61.25 & \textbf{62.09} & \textbf{71.98} & 40.68 & 51.67 \\
\textit{ Duration}         & 60.40 & \textbf{64.91} & 56.12 & 60.42 & \textbf{61.47} & 58.77 & 47.56 & 56.58 \\
\textit{ Frequency}        & 51.26 & \textbf{54.37} & 43.43 & 46.19 & \textbf{53.25} & 52.10 & 36.36 & 33.09 \\
\textit{ Combination.Drug} & \textbf{69.77} & \textbf{71.18} & 66.87 & 68.93 & 69.34 & 70.90 & 60.79 & 62.90 \\ \hline
Effect                                      & 74.33 & \textbf{84.73} & 74.68 & 83.94 & \textbf{74.75} & 84.65 & 64.60 & 79.19 \\ \hline
\end{tabular}}
\caption{Argument extraction results for each argument type. To accommodate space limitations, we showcase results for Flan-T5 with two augmentation strategies and ChatGPT. The Flan-T5 results represent the average score across 5-fold cross-validation, while the ChatGPT results showcase the performance of the 5-shot BM25 approach.}
\label{tab:rst-type}
\end{table*}

Table \ref{tab:rst-type} provides a detailed overview of argument extraction results for Flan-T5 with two augmentation strategies and ChatGPT. In comparison, ChatGPT exhibits a specific vulnerability in accurately matching main arguments, likely attributed to their greater length, which poses challenges in precise boundary determination. When it comes to sub-arguments, ChatGPT demonstrates a performance distribution similar to fine-tuning models but achieves lower overall scores. Notably, for certain argument types of which ChatGPT performs notably worse, such as `\emph{frequency}' and `\emph{duration}', these shortcomings also negatively impact the performance when training with ChatGPT-generated data. However, after filtering, the performance on these argument types can be improved to the extent that they may even outperform fine-tuning with annotated training data alone.

\section{Supplementary material on qualitative analysis for data augmentation }
\label{apd:qualitative-analysis}

To elucidate the potential performance decline associated with synthetic data, we sampled five instances for each argument type, where the fine-tuned model made correct predictions, while the augmented model (without filtering) made incorrect predictions. We conducted a statistical analysis of the error types and Table \ref{tab:data-aug-quality-arg} presents the distribution of error categories for each argument type.

To delve further into the origins of these errors, we also sampled the data generated by ChatGPT, conducting a statistical analysis of its labelling errors for comparison. Given the absence of a comparable gold standard for the synthesized data by ChatGPT, we randomly sampled 30 generated cases and assessed errors across all argument types. The resulting statistical findings are detailed in Table \ref{tab:syn-data-qual}.


\begin{table*}[htb]
\centering
\resizebox{0.8\textwidth}{!}{
\begin{tabular}{lcccc}
\hline
 &
\multicolumn{1}{c}{\begin{tabular}[c]{@{}c@{}}Argument\\ Missing\end{tabular}} &
  \multicolumn{1}{c}{\begin{tabular}[c]{@{}c@{}}Semantic \\ Misunderstanding\end{tabular}} &
  \multicolumn{1}{c}{\begin{tabular}[c]{@{}c@{}}Boundary\\ Problems\end{tabular}} &
  \multicolumn{1}{c}{\begin{tabular}[c]{@{}c@{}}Annotation \\ Noises\end{tabular}} \\ \hline
Subject & - & - & 2 & 3 \\
\textit{ Age}                & 3  & 2  & -  & - \\  
\textit{ Disorder}   & -  & 4  & 1  & - \\  
\textit{ Gender}             & 1  & 3  & -  & 1 \\  
\textit{ Population}         & 3  & -  & 2  & - \\  
\textit{ Race}               & -  & -  & -  & - \\ \hline
Treatment & - & 1 & 4 & - \\
\textit{ Drug}               & 1  & 1  & 1  & 2 \\  
\textit{ Treatment.Disorder} & 3  & 1  & 1  & - \\  
\textit{ Route}              & 3  & -  & 1  & 1 \\  
\textit{ Dosage}             & 1  & 1  & 3  & - \\  
\textit{ Time elapsed}       & 2  & 2  & 1  & - \\  
\textit{ Duration}           & 1  & 2  & 1  & 1 \\  
\textit{ Frequency}          & 1  & 3  & 1  & - \\  
\textit{ Combination.Drug}   & 1  & 4  & -  & - \\ \hline
Effect & - & 2 & 3 & - \\ \hline
\textbf{Total}                                         & 20 & 26 & 21 & 8 \\ \hline
\end{tabular}}
\caption{Statistics of error types for each argument type in qualitative analysis for data augmentation.}
\label{tab:data-aug-quality-arg}
\end{table*}

\begin{table*}[htb]
\centering
\resizebox{0.97\textwidth}{!}{
\begin{tabular}{lcccccc}
\hline
 &
  \multicolumn{1}{c}{\begin{tabular}[c]{@{}c@{}}Semantic\\ Misunderstanding\end{tabular}} &
  \multicolumn{1}{c}{\begin{tabular}[c]{@{}c@{}}Semantic\\ Incompleteness\end{tabular}} &
  \multicolumn{1}{c}{\begin{tabular}[c]{@{}c@{}}Argument\\ Missing\end{tabular}} &
  \multicolumn{1}{c}{\begin{tabular}[c]{@{}c@{}}Boundary\\ Problem\end{tabular}} &
  \multicolumn{1}{c}{\begin{tabular}[c]{@{}c@{}}\#. In Synthesized \\ Samples\end{tabular}} &
  \multicolumn{1}{c}{\begin{tabular}[c]{@{}c@{}}\#. In Template \\ Samples\end{tabular}} \\ \hline
Subject                                     & - & - & - & - & 17 & 15 \\
\textit{ Age}              & - & - & - & - & 2 & 7 \\
\textit{ Disorder}         & 1 & - & 2 & - & 8 & 10 \\
\textit{ Gender}           & - & - & - & - & 4 & 4 \\
\textit{ Population}       & - & - & - & - & - & - \\
\textit{ Race}             & - & - & - & - & - & -  \\ \hline
Treatment                  & - & 1 & - & - & 30 & 30 \\
\textit{ Drug}             & 1 & - & 5 & 2 & 32 & 36 \\
\textit{ Disorder}         & 1 & - & - & - & 8 & 11 \\
\textit{ Route}            & - & - & - & - & 2 & 6 \\
\textit{ Dosage}           & - & - & - & - & - & - \\
\textit{ Time elapsed}     & 1 & - & - & - & 1 & 1 \\
\textit{ Duration}        & - & - & - & - & 2 & 1 \\
\textit{ Frequency}        & - & - & - & - & - & 1 \\
\textit{ Combination.Drug} & 1 & - & 3 & - & 6 & 9 \\ \hline
Effect                     & - & 2 & - & 1 & 29 & 28 \\ \hline
\textbf{Total} & 5 & 3 & 5 & 3 & 141 & 159 \\ \hline
\end{tabular}}
\caption{Results of qualitative analysis for ChatGPT-synthesized data.}
\label{tab:syn-data-qual}
\end{table*}

\section{Prompt Details}
\label{apd:prompt}

Table \ref{tab:apd-prompt} shows the instructions utilized for ChatGPT's zero-shot prompting. Through our preliminary experiments, we discovered that ChatGPT exhibits better performance when tasked with generating structured output in JSON format rather than textual output. Based on this finding, we explore additional possibilities. For the end-to-end generation approach, we experiment with modifying the instructions to a code style or providing a detailed explanation of the schema. In the case of pipeline prompting, we initially prompt ChatGPT to generate the skeleton of the output, encompassing multiple events in a competent manner. Subsequently, in the second stage, we provide the generation from the first stage and ask specific questions for each sub-argument type.

Table \ref{tab:syn-prompt} presents the prompt employed to query ChatGPT for the generation of synthesized instances for examples with adverse events. We employ a similar prompt for the data generation of cases with potential therapeutic events and multiple events. Differently, we apply only the drug constraint to instances related to potential therapeutic events, as these typically do not involve a relevant effect. In addition, we refrain from imposing such constraints on multi-event instances, as doing so may complicate the preservation of event structure in synthesized samples.


\section{Licenses}

The PHEE dataset employed in this study is subject to the MIT License. The UIE model is covered by the Creative Commons Attribution-NonCommercial-ShareAlike 4.0 International Public License. The Flan-T5 model under the Apache License 2.0, and ChatGPT is a commercial service for which we adhere to OpenAI's terms of use. We use the dataset and tools within the scope of their intended use.

\clearpage
\onecolumn

\begin{xltabular}{0.9\textwidth}{p{0.2\textwidth}|p{0.75\textwidth}}

\toprule
Prompting Strategy & Example \\ \midrule
\textbf{Schema}& Extract event information from the following sentence and return events in json format as this: [\{"event\_type": event type, "arguments":[\{"argument\_type": argument type, "argument\_span":argument extraction\}]\}]. Event type: adverse event, potential therapeutic event. Argument type: subject, age, gender, race, population, subject\_disorder, treatment, drug, dosage, route, duration, frequency, time\_elapsed, indication, combination\_drug, effect. Sentence: <SENTENCE> Output: \\\midrule
\textbf{Code}& Argument = \{"argument\_type": str, \#options: [subject, age, gender,race, population, subject\_disorder, treatment, drug, dosage, route, duration, frequency, time\_elapsed, indication, combination\_drug, effect]\newline "argument\_span": str,\}\newline Event =\{"event\_type": str, \#options: [adverse\_event, potential\_therapeutic\_event]\newline "arguments": List[Argument],\}\newline events: List[Event] = extract events in the sentence: <SENTENCE>\newline print(json.dumps(events)) \\\midrule
\textbf{Explanation}& Extract event information from the following sentence and return events in json format as this: [\{"event\_type": event type, "arguments":[\{"argument\_type": argument type, "argument\_span":argument extraction\}]\}]. Event type: adverse event (an event shows the use of a drug or combination of drugs cause a harmful effect on the human patient), potential therapeutic event (an event shows the use of a drug or combination of drugs bring a potential beneficial effect on the human patient). Argument type: subject (overall description of the patients involved in the event), age (the concrete age or an age range of the subject), gender (the subject's gender), race (the subject's race or nationality), population (the number of patients receiving the treatment), subject\_disorder (the subject's disorders), treatment (overall description of the therapy administered to the patients), drug (the drugs used as therapy in the event), dosage (the amount of the drug is given), route (the route of the drug administration), duration (how long the patient has been taking the medicine), frequency (the frequency of drug use), time\_elapsed (the time elapsed after the drug was administered to the occurrence of the side effect), indication (the target disorder of the medicine administration), combination\_drug (the drugs used in combination), effect (the side effect in the adverse event or the beneficial effect in the potential therapeutic event). Sentence: <SENTENCE> Output:\\\midrule
\textbf{Pipeline}& \textbf{Stage 1}:\newline Extract adverse events and potential therapeutic events in the sentence, as well as the information about the subject (the patient), the treatment and the effect of the treatment involved in the event. Return the output in json format as this: [\{"event\_type": event type, "subject": span of subject information, "treatment": span of treatment information, "effect": span of effect information\}]. Event type: adverse event, potential therapeutic event. Sentence: <SENTENCE> Output: \newline \textbf{Stage 2}: Answer the question related to the given sentence and given event information. The answer should be a span exactly extracted from the sentence. If no answer can be found from the sentence, return N/A. Sentence: <SENTENCE> Event: Event type: <EVENT\_TYPE> Subject: <SUBJECT> Treatment: <TREATMENT> Effect: <EFFECT>. <QUESTION> \newline \textbf{Questions for each sub-argument type:}\newline
\textit{age}: What's the age of the subject?\newline
\textit{gender}: What's the gender of the subject?\newline
\textit{race}: What's the race or the nationality of the subject?\newline
\textit{population}: How many subjects are involved in the event?\newline
\textit{subject\_disorder}: What disorders do the subjects suffer from?\newline
\textit{drug}: What drugs are administered to the subject?\newline
\textit{dosage}: What amount of the drug is administered to the subject?\newline
\textit{route}: What route is the drug given to the subject?\newline
\textit{duration}: How long have the subject been taking the drug until the event occurred?\newline
\textit{frequency}: How frequently does the subject take the drug?\newline
\textit{time\_elapsed}: How long has elapsed since the patient started or ended dosing until the event occurred?\newline
\textit{indication}: What's the target disease of the treatment?\newline
\textit{combination\_drug}: What drugs are used in combination in the event\newline
\\
\bottomrule
\caption{Instructions for zero-shot prompting. <SENTENCE> is replaced with the query sentence. In the second stage of the pipeline prompting, <EVENT\_TYPE>, <SUBJECT>, <TREATMENT>, <EFFECT> are replaced with the generated results from the first stage, and <QUESTION> is replaced with manually crafted questions for each argument type. To enhance clarity, we substitute the `\textit{treatment\_disorder}' in the dataset with `\textit{indication}' when querying ChatGPT. }
\label{tab:apd-prompt}
\end{xltabular}
\clearpage
\twocolumn

\begin{table*}[t]
\centering
\begin{tabular}{p{0.95\textwidth}}
\toprule
Sentence: <SENTENCE> The events involved in the sentence are: <OUTPUT> Event type: adverse event (an event shows the use of a drug or combination of drugs cause a harmful effect on the human patient), potential therapeutic event (an event shows the use of a drug or combination of drugs bring a potential beneficial effect on the human patient). Argument type: subject (overall description of the patients involved in the event), age (the concrete age or an age range of the subject), gender (the subject's gender), race (the subject's race or nationality), population (the number of patients receiving the treatment), subject\_disorder (the subject's disorders), treatment (overall description of the therapy administered to the patients), drug (the drugs used as therapy in the event), dosage (the amount of the drug is given), route (the route of the drug administration), duration (how long the patient has been taking the medicine), frequency (the frequency of drug use), time\_elapsed (the time elapsed after the drug was administered to the occurrence of the side effect), indication (the target disorder of the medicine administration), combination\_drug (the drugs used in combination), effect (the side effect in the adverse event or the beneficial effect in the potential therapeutic event).  Generate a sentence with an adverse event which has a similar structure as the given sentence, and extract the events in the generated sentence. The drug <CONST\_DRUG> must appear in the event, and the effect should be <CONST\_EFFECT>. Return in the following json format: \{"sentence":the generated sentence, "output": [\{"event\_type": event type, "event\_trigger": the token indicating the existence of the event, "arguments":[\{"argument\_type": argument type, "argument\_span":argument extraction\}]\}]\}. Return the json output only. \\
\bottomrule
\end{tabular}
\caption{The prompt used to query ChatGPT for generating synthesized instances for ADE cases, with <SENTENCE> representing an example sentence from the training set, <OUTPUT> representing the annotation of the example sentence, <CONST\_DRUG> and <CONST\_EFFECT> representing a pair of sampled drug and effect from the training set.}
\label{tab:syn-prompt}
\end{table*}

\end{document}